\documentclass{article}

     \PassOptionsToPackage{numbers, compress}{natbib}

\usepackage[preprint,nonatbib]{neurips_2021}

\usepackage[utf8]{inputenc} 
\usepackage[T1]{fontenc}    

\usepackage{url}            
\usepackage{booktabs}       
\usepackage{amsfonts}       
\usepackage{nicefrac}       
\usepackage{microtype}      
\usepackage{xcolor}         
\usepackage{graphicx}
\usepackage{multirow}
\usepackage{amstext}
\usepackage{subfig}
\usepackage{amsmath}
\usepackage{adjustbox}

\usepackage[pagebackref=true,breaklinks=true,colorlinks,bookmarks=false]{hyperref}

\title{Local-to-Global Self-Attention in Vision Transformers}

\author{%
  Jinpeng Li\thanks{*Equal contribution} \\
  Inception Institute of \\ Artificial Intelligence (IIAI), \\ 
  Abu Dhabi, UAE\\
  \texttt{ljpadam@gmail.com} \\
  \And
  Yichao Yan\footnotemark[1] \footnotemark[2] \\
  MoE Key Lab of Artificial Intelligence,\\
  AI Institute,\\ Shanghai Jiao Tong University, China\\
  \texttt{yanyichao91@gmail.com}
  \AND
    Shengcai Liao\thanks{Corresponding author} \\
  Inception Institute of \\ Artificial Intelligence (IIAI), \\ Abu Dhabi, UAE\\
  \texttt{scliao@ieee.org } \\
  \And
  Xiaokang Yang \\
  MoE Key Lab of Artificial Intelligence,\\
  AI Institute,\\ Shanghai Jiao Tong University, China\\
  \texttt{xkyang@sjtu.edu.cn} \\
  \AND
  Ling Shao \\
  Inception Institute of Artificial Intelligence (IIAI), Abu Dhabi, UAE\\
  \texttt{ling.shao@ieee.org } \\
  \AND
  
}

\begin{document}

\maketitle

\begin{abstract}
  Transformers have demonstrated great potential in computer vision tasks. To avoid dense computations of self-attentions in high-resolution visual data, some recent Transformer models adopt a hierarchical design, where self-attentions are only computed within local windows. This design significantly improves the efficiency but lacks global feature reasoning in early stages. In this work, we design a multi-path structure of the Transformer, which enables local-to-global reasoning at multiple granularities in each stage. The proposed framework is computationally efficient and highly effective. With a marginal increasement in computational overhead, our model achieves notable improvements in both image classification and semantic segmentation. Code is available at \def\UrlFont{\em}\url{https://github.com/ljpadam/LG-Transformer}. 
\end{abstract}

\section{Introduction}
Over the past decade, we have witnessed the continuous success of convolutional neural networks (CNNs) in computer vision. Recently, Visual Transformers have also demonstrated strong potential, achieving state-of-the-art performance in several visual tasks, including image classification, object detection, and semantic segmentation. Transformers originally emerged from natural language processing (NLP) models and are considered as more flexible alternatives to CNNs for processing multi-modal inputs.

One of the major differences between a CNN and Transformer is the feature interaction mechanism. In a CNN, convolutional kernels are locally connected to the input feature maps, where features only interact with their local neighbors. In contrast, Transformer models, such as ViT~\cite{DBLP:journals/corr/abs-2010-11929}, require global feature reasoning by computing self-attentions among all the tokens. As a result, Transformers are computationally inefficient when processing images with a large number of visual tokens. To address this issue, several recent works, such as Swin Transformer~\cite{DBLP:journals/corr/abs-2103-14030}, CvT~\cite{DBLP:journals/corr/abs-2103-15808}, and PVT~\cite{DBLP:journals/corr/abs-2102-12122}, take inspiration from CNN models, and only compute self-attentions within local windows. This strategy brings significant improvements in efficiency, but abandons global feature reasoning in early stages, which weakens the potential of Transformer models.

In this work, we study the necessity of global and local feature interactions in self-attention modules of Vision Transformers and propose a local-to-global multi-path mechanism in self-attention, which we refer to as the LG-Transformer. Fig.~\ref{fig:arch} provide an outline of our architecture, which follows the macro-structure of locally connected Transformers~\cite{DBLP:journals/corr/abs-2103-14030,DBLP:journals/corr/abs-2103-15808,DBLP:journals/corr/abs-2102-12122}. In contrast to these models, in the early stages (stage 1-3), we expand the self-attention module into a multi-path structure, where feature maps (tokens) are downsampled so as to cover different local scales. To enable global reasoning, we set the window size equal to the smallest feature scale (i.e., $\frac{H}{32}\times \frac{W}{32}$). After the multi-path self-attention, the local-to-global features are aggregated as a more discriminative representation for the tokens. As the scales of global features are significantly downsampled at each stage, the computational overhead is marginally increased compared with the single-path locally connected baseline Transformer. In the meantime, the multi-path self-attention structure is easy to implement, and it does not bring in novel operations, which yields an efficient and strong Transformer model.

\begin{figure}
  \centering
  \subfloat[Multi-granular conv in the Inception Network~\cite{DBLP:conf/cvpr/SzegedyLJSRAEVR15}\label{subfig-2-1}]{%
   \includegraphics[width=0.47\linewidth]{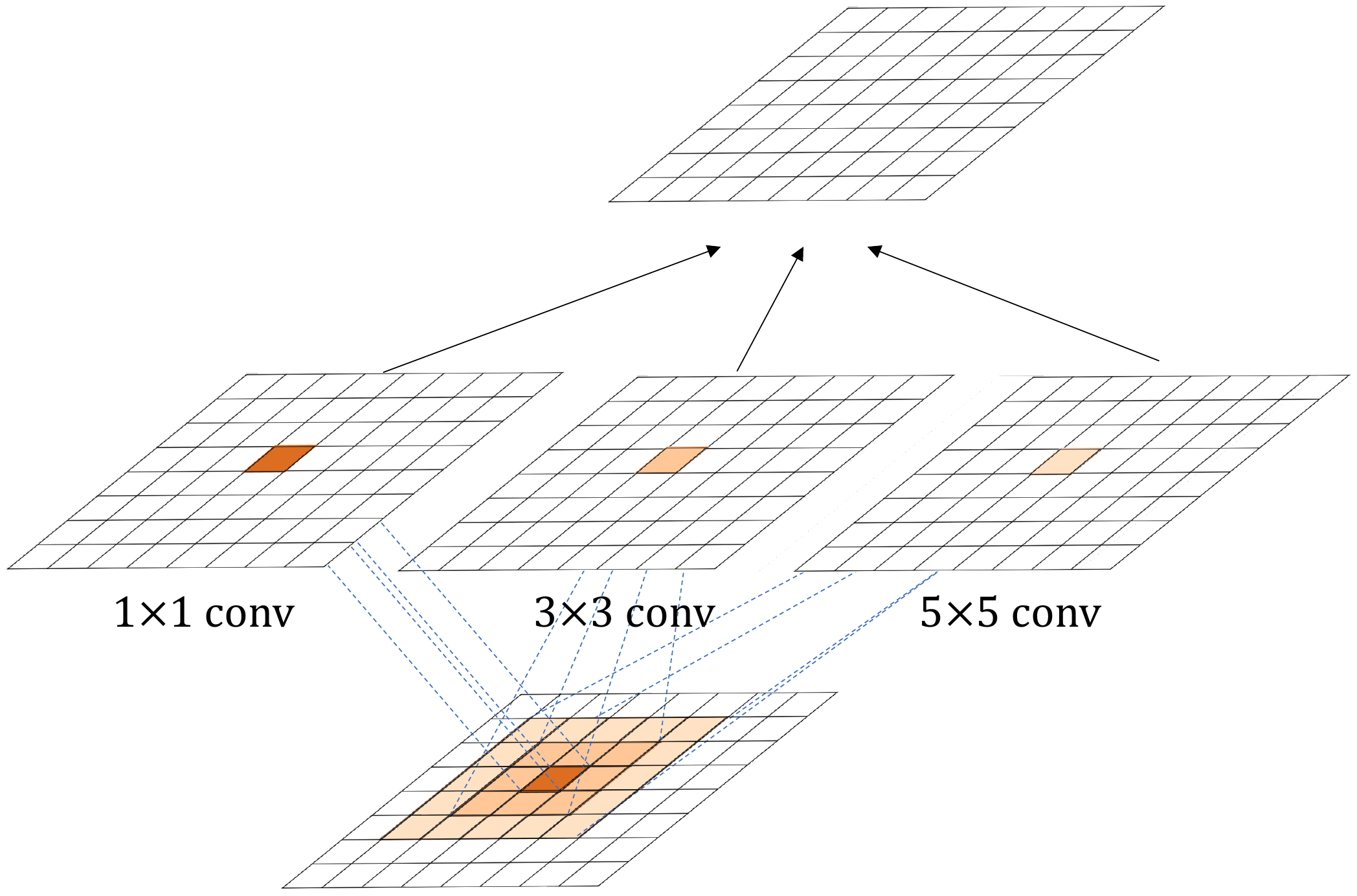}
}
\hspace{4mm}
\subfloat[Local-to-Global self-Attention in our framework\label{subfig-2-2}]{%
   \includegraphics[width=0.47\linewidth]{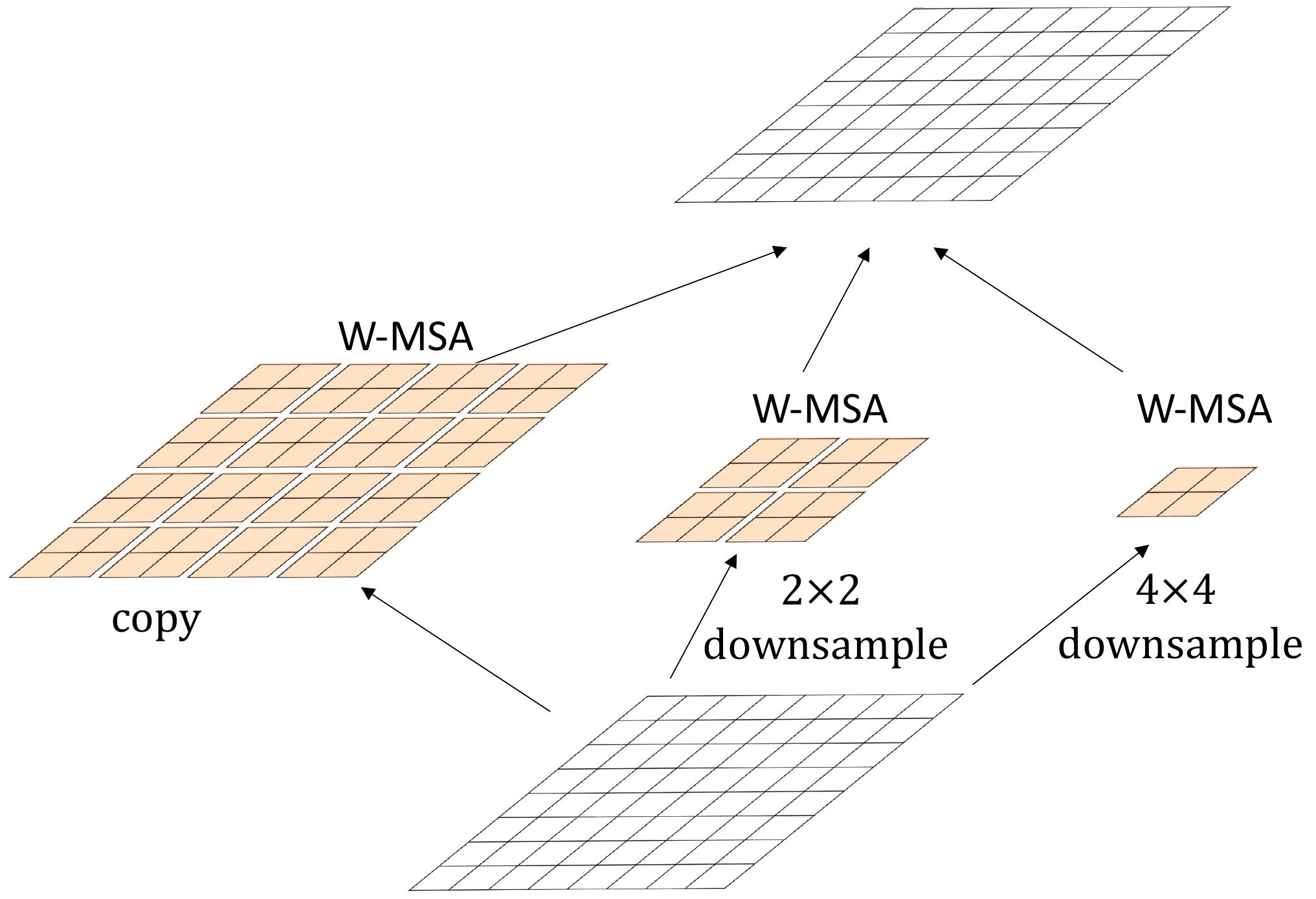}
}
  \caption{Our design is inspired by the multi-granular convolutional structure in CNN models, where $1\times1$, $3\times3$ and $5\times5$ convolution kernels captures different levels of local information. In our framework, feature maps are down-sampled into different scales, and window-based multi-head self attention (W-MSA) is applied to each scale to model local-to-global feature interaction. }
  \label{fig:intro}
\end{figure}

Our design is inspired by the multi-granular convolutional structure in CNN models, such as Inception Networks~\cite{DBLP:conf/cvpr/SzegedyLJSRAEVR15,DBLP:conf/cvpr/SzegedyVISW16,DBLP:conf/aaai/SzegedyIVA17} and High Resolution Networks~\cite{DBLP:conf/cvpr/0009XLW19}, where feature interactions are computed within local regions of different scales. Different from these CNN frameworks, our model not only computes the local feature interactions, but also enables global feature reasoning, taking advantage of both CNN and Transformer models. A comparison is illustrated in Fig.~\ref{fig:intro}.

LG-Transformer achieves competitive results on both image classification and semantic segmentation. Compared with the state-of-the-art model Swin-Transformer~\cite{DBLP:journals/corr/abs-2103-14030}, LG-Transformer improves the Top-1 accuracy by 0.9\% on ImageNet~\cite{DBLP:conf/cvpr/DengDSLL009} classification. Meanwhile, it improves the mIoU by 0.8\% on ADE20K~\cite{DBLP:conf/eccv/LinMBHPRDZ14} for semantic segmentation.

\section{Related Work}
\textbf{Transformers and Vision Transformers}.
The pioneering deep neural network architecture, Transformer~\cite{DBLP:conf/nips/VaswaniSPUJGKP17}, was firstly proposed to solve sequence transduction tasks in NLP. The original Transformer is built on self-attention modules which excel at modeling long-range information under arbitrary input lengths. Based on this simple and powerful architecture, several Transformer variants~\cite{DBLP:conf/naacl/DevlinCLT19,DBLP:conf/nips/BrownMRSKDNSSAA20,DBLP:journals/corr/abs-1907-11692,DBLP:conf/emnlp/LiuL19} achieve dominant positions in many NLP tasks, such as machine translation and summarization. 

Motivated by their great success in NLP, self-attention mechanisms and Transformer have also been introduced to computer vision~\cite{DBLP:conf/eccv/CarionMSUKZ20,DBLP:journals/corr/abs-2010-04159,DBLP:conf/cvpr/YangYFLG20,DBLP:conf/iccv/SunMV0S19,DBLP:journals/corr/abs-2012-00364}. The early Transformer-based vision models mainly leverage the advantage of global dependencies. For example, the Vision Transformer (ViT)~\cite{DBLP:journals/corr/abs-2010-11929} is a pure Transformer-based image classification model which takes image patches as tokens and directly models the global attention on all the tokens. It achieves comparable performance as hand-designed and automatically searched CNN architectures~\cite{DBLP:conf/cvpr/HeZRS16,DBLP:journals/corr/abs-2012-00364}. However, due to the large resource cost of global attention, ViT is more suitable to generate low-resolution outputs, which limits its applications. 

To address this issue, a Pyramid Vision Transformer (PVT)~\cite{DBLP:journals/corr/abs-2102-12122} is proposed to reduce the resolutions of keys and values with a spatial-reduction attention, which improves the efficiency while still maintaining the capability of aggregating global information. Although global connectivity in Transformer models shows promising results in computer vision, it is undeniable that existing CNN models also demonstrate the importance of local information. Some works try to combine the advantages of CNN and self-attention by appended blocks~\cite{DBLP:journals/corr/abs-2006-03677}, interleave architecture~\cite{DBLP:conf/iccv/BelloZLVS19,DBLP:journals/corr/abs-2104-05707} and parallel paths~\cite{DBLP:journals/corr/abs-2012-00759}. LocalViT~\cite{DBLP:journals/corr/abs-2104-05707} employs depth-wise convolution layers into transformers to explicitly model the local dependencies. CvT~\cite{DBLP:journals/corr/abs-2103-15808} introduces convolution into token embeddings and feature projection to combine local information and global image contexts. Swin-Transformer~\cite{DBLP:journals/corr/abs-2103-14030} explores local information by a pure transformer based architecture which splits feature maps into non-overlapping windows, and applies intra-window and cross-window interactions to build hierarchical attentions.

\textbf{Multi-granular connection in CNN}. Multi-granular information has been widely explored in computer vision since the era of hand-crafted features~\cite{DBLP:journals/ijcv/Lowe04,DBLP:conf/cvpr/DalalT05,DBLP:journals/pami/FelzenszwalbGMR10,DBLP:journals/pami/DollarABP14}. Although CNN models~\cite{DBLP:conf/cvpr/HeZRS16,DBLP:journals/corr/abs-2012-00364,DBLP:journals/corr/SimonyanZ14a,huang2017densely} demonstrate strong feature representation capability due to the end-to-end learning and deep hierarchical architectures, they still benefit from multi-granular connections in many computer vision applications, such as image classification~\cite{DBLP:conf/nips/SaxenaV16,DBLP:conf/cvpr/SzegedyLJSRAEVR15}, object detection~\cite{DBLP:conf/eccv/CaiFFV16} and semantic segmentation~\cite{DBLP:journals/pami/ChenPKMY18,DBLP:conf/bmvc/FourureEFMT017}. Inception Networks~\cite{DBLP:conf/cvpr/SzegedyLJSRAEVR15,DBLP:conf/cvpr/SzegedyVISW16,DBLP:conf/aaai/SzegedyIVA17} are a serious of classification models built upon multi-granular convolution kernels to improve the classification performance and computation efficiency. HRNet~\cite{DBLP:conf/cvpr/0009XLW19,DBLP:conf/eccv/YuanCW20} constructs multi-granular feature maps in parallel paths to simultaneously maintain more semantic and spatial information, and it achieves promising performance on several dense prediction tasks. To detect objects with various sizes, pyramid feature maps are generated by connecting bottom-up and top-down pathways in many CNN based detectors~\cite{DBLP:conf/cvpr/LinDGHHB17,DBLP:conf/eccv/ZhuDHFXQH18,DBLP:conf/cvpr/GhiasiLL19}. PSPNet~\cite{DBLP:conf/cvpr/ZhaoSQWJ17} introduces a pyramid pooling module to fuse multi-granular context information for semantic segmentation methods. Hourglass~\cite{DBLP:conf/eccv/NewellYD16} stacks multiple encoder and decoder networks and builds multi-granular connections among them for human pose estimation.

In this work, we take advantage of both the local connectivity in CNNs and global connectivity in Transformers, building a Transformer model with local-to-global attention, yielding a framework with improved performance and meanwhile maintains high efficiency.

\section{Method}
\begin{figure}
  \centering
  \includegraphics[width=\linewidth]{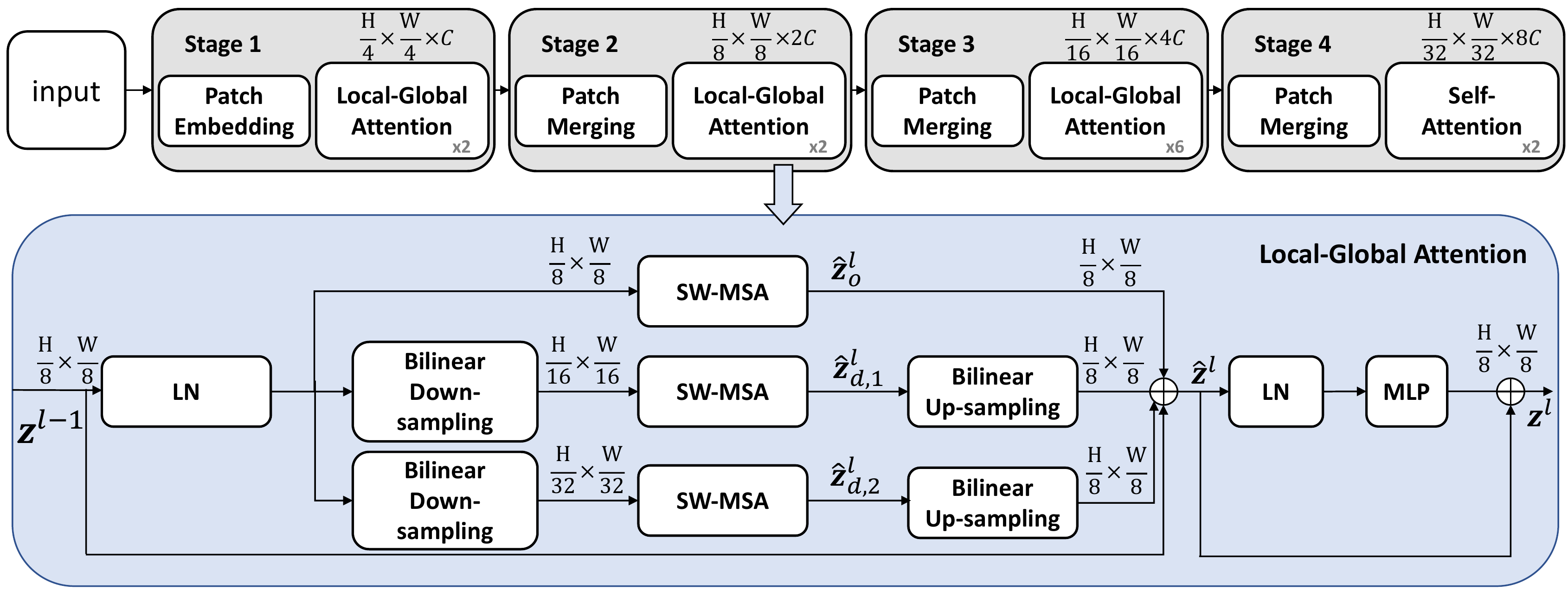}
  \caption{Our Local-to-Global (LG) Transformer contains four stages, each of which includes a patch embedding/merging layer, followed by several LG-Attention blocks. Each LG-Attention block contains several parallel SW-MSA modules to calculate local attentions, gathering local-to-global information with feature interactions.}
  \label{fig:arch}
\end{figure}

In this section, we describe the detailed architecture of the proposed LG-Transformer. As illustrated in Fig.~\ref{fig:arch}, our model follows the hierarchical design of several recent visual Transformers~\cite{DBLP:journals/corr/abs-2103-14030,DBLP:journals/corr/abs-2103-15808,DBLP:journals/corr/abs-2102-12122} for efficient learning. Our framework contains four stages. In the first stage, patch embeddings are applied to the input image, resulting in $\frac{H}{4}\times\frac{W}{4}$ tokens. In each of the subsequent stages, the resolution of feature maps (tokens) is reduced by a factor of 4 ($2\times2$) with a patch merging layer~\cite{DBLP:journals/corr/abs-2103-14030}, while the feature dimensions are increased by a factor of 2. Finally, the network outputs $\frac{H}{32}\times\frac{W}{32}$ visual tokens.

\subsection{Local-to-Global Attention Block}
The core design of our framework is the Local-to-Global (LG) Attention block, applied to stages 1-3. As illustrated in the bottom part of Fig.~\ref{fig:arch}, the LG-Attention block contains several parallel attention paths. Take stage 2 for example. After layer normalization (LN), the input feature $z^1$ with resolution $\frac{H}{8}\times\frac{W}{8}$ is bilinearly down-sampled into feature maps with lower dimensions (\emph{i.e.}, $\frac{H}{16}\times\frac{W}{16}$ and $\frac{H}{32}\times\frac{W}{32}$). Each feature map is processed with a window based multi-head self-attention with shifted window partitioning (SW-MSA)~\cite{DBLP:journals/corr/abs-2103-14030}. Specifically, feature maps are partitioned into non-overlapped windows, and self-attentions are only calculated within each window, thus the computation complexity is significantly reduced compared performing self-attention on the entire feature map. By displacing the window between consecutive attention blocks, SW-MSA enables local patches to communicate with its different neighbors. However, the feature interaction is still limited to a local region. In this work, we set the window size to $[\frac{H}{32},\frac{W}{32}]$, such that the feature map with the lowest resolution can be processed with global self-attention. After the SW-MSA modules, the downsampled features maps are upsampled to the same resolution and combined, before being fed into the LN and MLP layers. The LG-Attention block can be calculated as:

\begin{equation}
	{{\hat{\bf{z}}}^{l}_{o}} = \text{SW-MSA}( {\text{LN}( {{{\bf{z}}^{l - 1}}} )} ) , 
\end{equation}
\begin{equation}
	{{\hat{\bf{z}}}^{l}_{d,1}} = \text{SW-MSA}( \text{BD}_{1}( {{\rm LN}(  {{{\bf{z}}^{l - 1}}} )}) ) , 
\end{equation}
\begin{equation}
	{{\hat{\bf{z}}}^{l}_{d,2}} = \text{SW-MSA}( \text{BD}_{2}({{\rm LN}(  {{{\bf{z}}^{l - 1}}} )}) ) , 
\end{equation}
\begin{equation}
    {\hat{\bf{z}}}^{l} =
    {\hat{\bf{z}}}_o^{l} + 
    \text{BU}_1({\hat{\bf{z}}}^{l}_{d,1}) + 
    \text{BU}_2({\hat{\bf{z}}}^{l}_{d,2}) +
    \bf{z}^{l-1},
\end{equation}
\begin{equation}
	{{\bf{z}}^l} = \text{MLP}( {\text{LN}( {{{\hat{\bf{z}}}^{l}}} })) + {{\hat{\bf{z}}}^{l}},
\end{equation}
where BD and BU denote bilinear downsampling and bilinear upsampling, respectively. ${\bf{z}}^{l-1}$ is the input feature,  ${\hat{\bf{z}}}^{l}_{o}$, ${\hat{\bf{z}}}^{l}_{d,1}$ and ${\hat{\bf{z}}}^{l}_{d,2}$ are the intermediate features containing local-to-global information, ${\hat{\bf{z}}}^{l}$ is the aggregation of these features, and ${\bf{z}}^{l}$ is the output feature.

\subsection{Computational Complexity}
Due to the downsampling operations in the LG-Attention blocks, the computational complexity is marginally increased compared with the Swin-Transformer baseline. Suppose each local window contains $M \times M$ tokens. The complexities of a SW-MSA module and our LG-Attention block (with parallel down-sampled ratios of 2 and 4) based on a feature map containing $h\times w$ tokens are\footnote{We omit the computation of Softmax and window shift.}:

\begin{equation}
    \Omega(\text{SW-MSA}) = 4hwC^2 + 2M^2hwC,
\end{equation}
\begin{equation}
    \Omega(\text{LG-Att}) = 5.25hwC^2 + 2.625M^2hwC.
\end{equation}
Although two novel branches are added to the network, LG-Attention block only increases the computations about 0.3 times compared with SW-MSA. Further, the complexity is a linear function of $hw$, making it significantly lower than the complexity of the global self-attention, which is proportional to $(hw)^2$.

\begin{table*}[!t]
\small
\begin{center}
\begin{adjustbox}{width=1\textwidth}
\begin{tabular}{c|c|c|c|c}
\toprule
 & \begin{tabular}[c]{@{}c@{}}Downsp. Rate \\ (output size)\end{tabular} & Layer Name & LG-T  & LG-S \\
\hline
\hline
\multirow{3}{*}{stage 1} & \multirow{3}{*}{\begin{tabular}[c]{@{}c@{}}4$\times$\\ (56$\times$56)\end{tabular}} &Patch Embeding & C=4$\times$4, D=96, LN  & C=4$\times$4, D=96, LN \\
\cline{3-5}
& & LG-Att & $\begin{bmatrix}\text{S=[4$\times$, 16$\times$, 32$\times$]}\\\text{W=7$\times$7, D=96, H=3}\end{bmatrix}$ $\times$ 2   & $\begin{bmatrix}\text{S=[4$\times$, 16$\times$, 32$\times$]}\\\text{W=7$\times$7, D=96, H=3}\end{bmatrix}$ $\times$ 2 \\
\hline
\multirow{3}{*}{stage 2}  & \multirow{3}{*}{\begin{tabular}[c]{@{}c@{}}8$\times$\\ (28$\times$28)\end{tabular}} &Patch Merging & C=2$\times$2, D=192, LN & C=2$\times$2, D=192, LN  \\
\cline{3-5}
& & LG-Att & $\begin{bmatrix}\text{S=[8$\times$, 16$\times$, 32$\times$]}\\\text{W=7$\times$7, D=192, H=6}\end{bmatrix}$ $\times$ 2  & $\begin{bmatrix}\text{S=[8$\times$, 16$\times$, 32$\times$]}\\\text{W=7$\times$7, D=192, H=6}\end{bmatrix}$ $\times$ 2 \\
\hline
\multirow{3}{*}{stage 3}  & \multirow{3}{*}{\begin{tabular}[c]{@{}c@{}}16$\times$\\ (14$\times$14)\end{tabular}} &Patch Merging & C=2$\times$2, D=384, LN & C=2$\times$2, D=384, LN \\
\cline{3-5}
& & LG-Att & $\begin{bmatrix}\text{S=[16$\times$, 32$\times$]}\\\text{W=7$\times$7, D=384, H=12}\end{bmatrix}$ $\times$ 6 & $\begin{bmatrix}\text{S=[16$\times$, 32$\times$]}\\\text{W=7$\times$7, D=384, H=12}\end{bmatrix}$ $\times$ 18  \\
\hline
\multirow{3}{*}{stage 4} & \multirow{3}{*}{\begin{tabular}[c]{@{}c@{}}32$\times$\\ (7$\times$7)\end{tabular}} &Patch Merging & C=2$\times$2, D=768, LN & C=2$\times$2, D=768, LN \\
\cline{3-5}
& & SW-MSA & $\begin{bmatrix}\text{S=[32$\times$, ]}\\\text{W=7$\times$7, D=768, H=24}\end{bmatrix}$ $\times$ 2 & $\begin{bmatrix}\text{S=[32$\times$, ]}\\\text{W=7$\times$7, D=768, H=24}\end{bmatrix}$ $\times$ 2 \\
\bottomrule
\end{tabular}
\end{adjustbox}
\end{center}
\caption{Detailed architecture of LG-T and LG-S. C denotes the concatenation rate. D is the number of dimensions of the embedding. S are the scale rates of parallel attention paths in the LG-Attention block. W denotes the window size of SW-MSA. H is the number of heads in the multi-head attention.}
\label{table:arch}
\end{table*}

\subsection{Architecture Variants}
We build two variants of our framework, denoted as LG-T and LG-S, where the only difference is the number of LG-Attention blocks in each stage. The detailed architecture of our framework is reported in Table~\ref{table:arch}. Except for the LG-Attention blocks, LG-Transformer basically employs the same settings as Swin-Transformer~\cite{DBLP:journals/corr/abs-2103-14030}.

\begin{table*}[!t]
    \small
    \begin{center}
        \begin{tabular}{l|r|c|c|c|c}
            \toprule
             Network & Image Size & Params (M) & FLOPs (G) & Top-1 Acc. (\%) & Top-5 Acc. (\%) \\
             \midrule
             \multicolumn{4}{c}{CNN Based Models} \\ \midrule
             MobileNetV1~\cite{DBLP:journals/corr/HowardZCKWWAA17}    & $224 \times 224$ & 4.2 & 0.6 & 70.6 & -- \\
             MobileNetV2 (1.4)~\cite{DBLP:conf/cvpr/SandlerHZZC18} & $224 \times 224$ & 6.9 & 0.6 & 74.7 & -- \\
             ResNet-18~\cite{DBLP:conf/cvpr/HeZRS16}   & $224 \times 224$ & 11.7 & 1.8 & 69.8 & 89.1 \\
             ResNet-50~\cite{DBLP:conf/cvpr/HeZRS16}   & $224 \times 224$ & 25.6 & 4.1 & 76.1 & 92.9 \\
             DenseNet-169~\cite{huang2017densely}       & $224 \times 224$ & 14.2 & 3.4 & 75.6 & 92.8\\
             RegNet-4GF~\cite{DBLP:conf/cvpr/RadosavovicKGHD20} & $224 \times 224$ & 20.7 & 4.0 & 80.0 & -- \\
             RegNet-16GF~\cite{DBLP:conf/cvpr/RadosavovicKGHD20} & $224 \times 224$ & 84 & 16 & 82.9 & -- \\
             EfficientNet-B4~\cite{DBLP:conf/icml/TanL19} & $380 \times 380$ & 19.3 & 4.5 & 82.9 & 96.4 \\ 
             EfficientNet-B6~\cite{DBLP:conf/icml/TanL19} & $528 \times 528$ & 43 & 19 & 84.0 & 96.8 \\\midrule
             \multicolumn{4}{c}{Transformers Based Models} \\ \midrule
             DeiT-T~\cite{DBLP:journals/corr/abs-2012-12877}  & $224 \times 224$  &5.7 & 1.3 &72.2 & 91.1\\
             DeiT-S~\cite{DBLP:journals/corr/abs-2012-12877}  & $224 \times 224$ & 22.1 & 4.6 &79.8 &95.1\\
             CrossViT-S ~\cite{DBLP:journals/corr/abs-2103-14899} & $224 \times 224$ & 26.7 & 5.6&81.0 &--\\
    		T2T-ViT-14 ~\cite{DBLP:journals/corr/abs-2101-11986} & $224 \times 224$ &22& 5.2&81.5&--\\
    		TNT-S ~\cite{han2021transformer} & $224 \times 224$ & 23.8& 5.2&81.3&--\\
    		CoaT Mini ~\cite{DBLP:journals/corr/abs-2104-06399}& $224 \times 224$ &10&6.8&80.8&--\\
    		CoaT-Lite Small ~\cite{DBLP:journals/corr/abs-2104-06399} & $224 \times 224$ & 20 & 4.0&81.9&--\\
    		PVT-Small \cite{DBLP:journals/corr/abs-2102-12122}& $224 \times 224$  & 24.5 & 3.8 & 79.8&-- \\ 
    		CPVT-Small-GAP \cite{chu2021conditional}& $224 \times 224$  & 23 & 4.6& 81.5&-- \\
    		Swin-T \cite{DBLP:journals/corr/abs-2103-14030}& $224 \times 224$  & 29 & 4.5 &81.3&-- \\
    		LG-T (\textbf{ours}) & $224 \times 224$  &32.6 & 4.8 &\textbf{82.1} &\textbf{95.8}\\ \hline
        
             T2T-ViT-19 \cite{DBLP:journals/corr/abs-2101-11986}& $224 \times 224$ & 39.2& 8.9 &81.9&--\\
		    PVT-Medium \cite{DBLP:journals/corr/abs-2102-12122}& $224 \times 224$ & 44.2 & 6.7 & 81.2&--\\
		    Swin-S~\cite{DBLP:journals/corr/abs-2103-14030}&$224 \times 224$ & 50 & 8.7 &83.0&--\\
		    Swin-B~\cite{DBLP:journals/corr/abs-2103-14030}&$224 \times 224$ & 88 & 15.4 &\textbf{83.3}&--\\
		    LG-S (\textbf{ours}) & $224 \times 224$  &61.0 & 9.4 &\textbf{83.3} &\textbf{96.2}\\
             \bottomrule
             
        \end{tabular}
    \end{center}
    \vspace{-0.2cm}
    \caption{Image classification results on ImageNet datasets.}
    \label{tab:ic}
\end{table*}

\begin{table*}[!t]
    \begin{center}
        \begin{tabular}{l|c|c|c}
            \toprule
            Stages with LG-Att & Params (M) & FLOPs (G) & Top-1 Acc. (\%)\\ \midrule
             -- & 4.5& 28.3 & 78.44 \\ 
             1   & 4.5 & 28.4 & 78.68 \\ 
             1$\sim$2 & 4.6 & 29.0 & 78.89 \\ 
             1$\sim$3 & 4.8 & 32.6 & 79.72 \\ \bottomrule
        \end{tabular}
    \end{center}
    \caption{Ablation study by inserting LG Attention blocks into different stages.}
    \label{tbl:placement}
    \vspace{0cm}
\end{table*}

\section{Experimental Results}\label{sec:exp}
In this section, we report the experimental results of our LG-Transformer on two computer vision tasks, \emph{i.e.}, image classification, and semantic segmentation. Comparisons with state-of-the-art models are given in each subsection. We also provide a thorough ablation study on model components and structure variations.

\subsection{Image Classification}

\textbf{Experiment Settings}.
To achieve fair comparison with previous works, most of our experiment settings follows Swin-Transformer~\cite{DBLP:journals/corr/abs-2103-14030}. Experiments on image classification are conducted on the ImageNet-1K dataset~\cite{DBLP:conf/cvpr/DengDSLL009} which contains 1,000 classes, and 1.28M and 50K images for the training set and validation set, respectively. Resolutions of input images are set to $224\times224$ for both training and evaluation. AdamW with $\beta_1=0.9$ and $\beta_2=0.999$ is use as optimizer for all of our models. We train LG-T and LG-S for 300 epochs in the comparison with state-of-the-art methods, and 100 epochs in the ablation studies. The initial learning rate is 0.001 with a cosine scheduler for learning rate decay, and a linear warm-up with 20 epochs is employ to stabilize training. All models are trained on a cluster with 8 V100 GPUs. We set Batch size to 128 on each GPU, and it takes near 4 days and 6 days to train a LG-T and a LG-S with 300 epochs, respectively. Data augmentation is used in the training phase, and we employ the same parameters as in~\cite{DBLP:journals/corr/abs-2103-14030}. 
We evaluate the test images based on a center cropped image without any test-time augmentation.

\begin{figure}
  \centering
  \includegraphics[width=\linewidth]{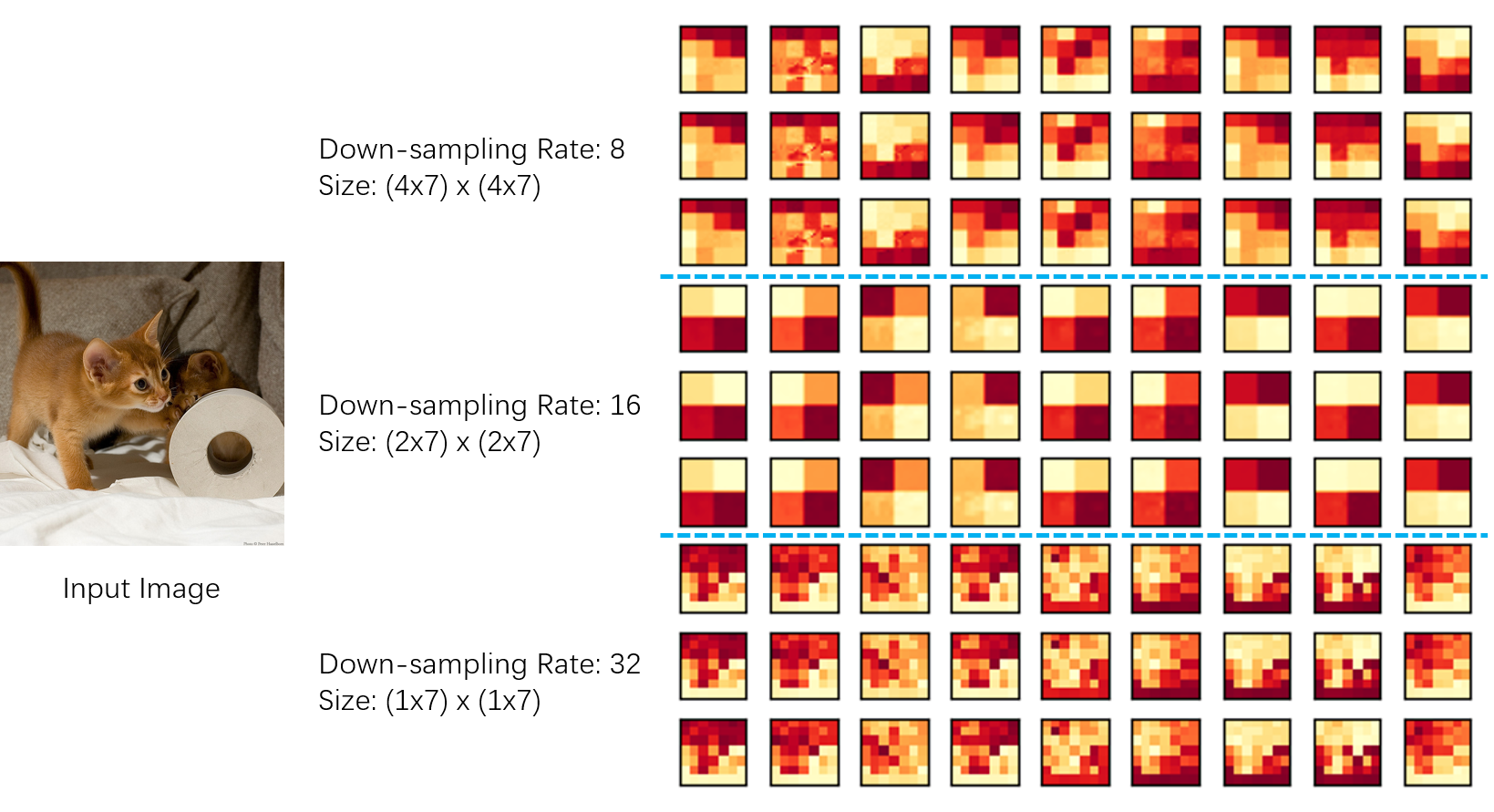}
  \caption{Visualization of feature maps in stage 2 of LG-T. The LG-Attention block contains three paths, which generate feature maps of the input image with downsampling rates of 8 (local attention), 16 (mid-level attention), and 32 (global attention).  Zoom in for better visualization.}
  \label{fig:feature_vis}
\end{figure}

\textbf{Comparative Results}. 
Table~\ref{tab:ic} presents the comparative results on ImageNet classification. When comparing LG-Transformer with state-of-the-art Vision Transformers, we observe that our framework achieves competitive or better results under similar computational overheads. For example, LG-T achieves 82.1\% top-1 accuracy, outperforming Swin-T by 0.8\%, PVT-Small by 2.3\%, T2T-ViT-14 by 0.6\%, and DeiT-S by 2.3\%. Meanwhile, LG-S also outperforms its Transformer counterparts, \emph{e.g.}, T2T-ViT-19, PVT-Medium, Swin-S. Notably, LG-S achieves comparable performance to Swin-B, with significantly fewer parameters and lower FLOPs. 

Further, the performance of LG-Transformer is significantly better than the widely utilized CNN models, \emph{i.e.}, MobileNet, ResNet, and DenseNet. Notably, the LG-Transformers achieve comparable performance to the state-of-the-art CNN models, \emph{i.e.}, RegNet and EfficientNet.

\begin{table*}[ht]
    \begin{center}
        \begin{tabular}{c|c|c|c|c}
            \toprule
            Scale $\frac{1}{16}$&Scale$\frac{1}{32}$ & Params (M) & FLOPs (G) & Top-1 Acc. (\%)\\ \midrule
              & & 4.5& 28.3 & 78.44 \\ 
             \checkmark& & 4.6 &  28.7&  78.74  \\ 
             &\checkmark & 4.7 & 32.2 & 79.60   \\ 
             \checkmark& \checkmark & 4.8 & 32.6 & 79.72 \\ \bottomrule
        \end{tabular}
    \end{center}
    \caption{Ablation study for LG-Attention paths.}
    \label{tbl:pth}
    \vspace{0cm}
\end{table*}

\begin{figure}
  \centering
  \subfloat[Training loss of LG-T and Swin-T.]{%
   \includegraphics[width=0.47\linewidth]{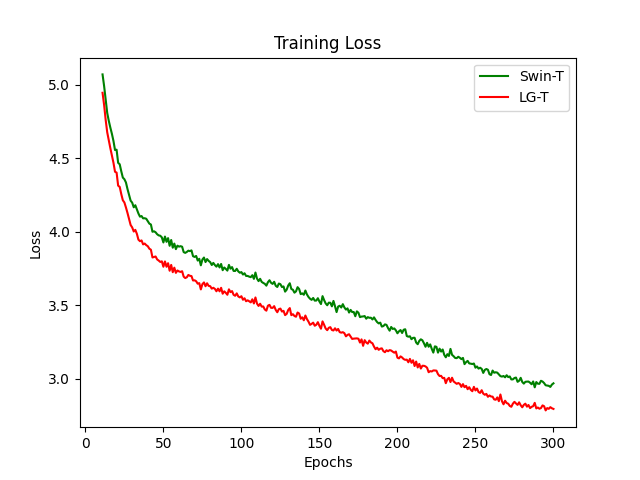}
}\quad
\subfloat[Validation accuracy of LG-T and Swin-T.]{%
   \includegraphics[width=0.47\linewidth]{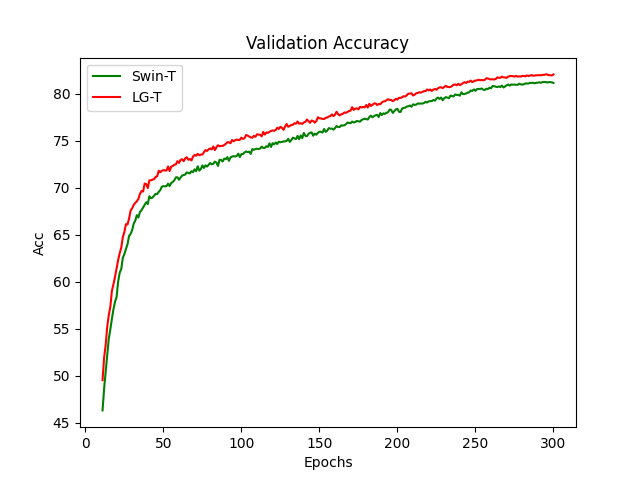}
}\\
\subfloat[Training loss of LG-T with different numbers of LG-Att.]{%
   \includegraphics[width=0.47\linewidth]{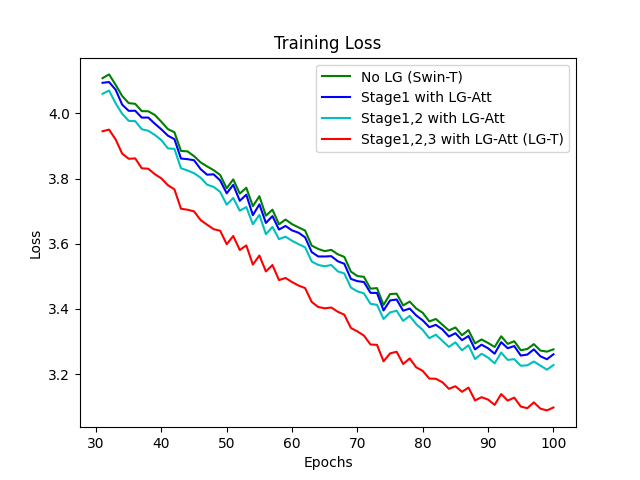}
}\quad
\subfloat[Validation accuracy of LG-T with different numbers of LG-Att.]{%
   \includegraphics[width=0.47\linewidth]{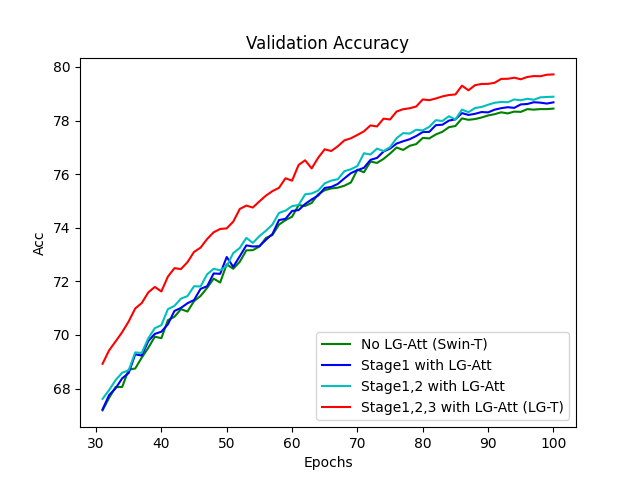}
}\\
\hspace{4mm}
  \caption{Training losses and validation accuracies on ImageNet.}
  \label{fig:loss}
\end{figure}

\textbf{Ablation Studies}. 
To evaluate the effectiveness of the proposed LG-Attention block, we carry out a series of ablation studies.\footnote{We report results by training 100 epochs.} First, as shown in Table~\ref{tbl:placement}, by inserting the LG-Attention block into different stages, we can observe that the performance of LG-T improves continuously (from 78.4\% to 79.7\%) when LG-Attention is employed in more stages, while the parameters and FLOPs are only marginally increased (from 28.3 GFLOPs to 32.6 GFLOPs).

Second, we compare different combinations of the attention paths with downsampling rates 16 and 32. As illustrated in Table~\ref{tbl:pth}, LG-Attention achieves the best performance when combining both types of attention. 

Furthermore, Fig.~\ref{fig:loss} shows the training losses and validation accuracies of different models. From Fig.~\ref{fig:loss}(a)(b), we find that GL-T converges faster, while consistently achieving a lower training loss and higher validation accuracy. When comparing the loss curves and validation accuracies of the model variants in Fig.~\ref{fig:loss}(c)(d), we find that adding LG-Attention blocks to more stages constantly improves the validation performance and training convergence.

In summary, these results validate the importance of the proposed multi-path attention mechanism and demonstrate the necessity of individual attention paths.

\textbf{Qualitative Results}. 
To better illustrate how the local-to-global attention blocks work, we provide some qualitative results containing the feature maps from stage 2 of LG-T in Fig.~\ref{fig:feature_vis}. We show three levels of feature maps, with downsampling rates of 8 (local attention), 16 (mid-level attention), and 32 (global attention). The basic building block of the feature maps are $7\times7$ feature blocks. For example, the feature maps with a downsampling rate of 8 contains $4\times4$ feature blocks. Within each block, only the $7\times7$ local features interact with each other. Feature maps with a downsampling rate of 16 contain 4 blocks of 7$\time7$ windows. Although the features in each block can represent more local information, each block is still processed individually. Attentions in the global feature maps a downsampling rate of 32 can cover the whole image, thus containing global semantic information. Notably, there is clear discontinuity across the borders of the local attention windows in the feature maps of downsampling rates of 8 and 16, showing the limitation of local attentions.

\begin{table*}[t]
    \begin{center}
\begin{tabular}{cc|c|cc}
\hline
\multicolumn{2}{c|}{ADE20K}& val & \multirow{2}{*}{\#param.} & \multirow{2}{*}{FLOPs} \\
Method & Backbone & mIoU  &  &  \\
\hline
DANet~\cite{DBLP:conf/cvpr/FuLT0BFL19} & ResNet-101 & 45.2  & 69M & 1119G  \\
ACNet~\cite{DBLP:conf/iccv/FuLW0BTL19} & ResNet-101 & 45.9 & - & - \\
DNL~\cite{DBLP:conf/eccv/YinYCLZLH20} & ResNet-101 & 46.0 & 69M & 1249G  \\
OCRNet~\cite{DBLP:conf/eccv/YuanCW20} & ResNet-101 & 45.3 & 56M & 923G \\
OCRNet~\cite{DBLP:conf/eccv/YuanCW20} & HRNet-w48 & 45.7 &71M & 664G \\
DeepLab.v3+~\cite{DBLP:conf/eccv/ChenZPSA18} & ResNet-101 & 44.1 & 63M & 1021G \\
DeepLab.v3+~\cite{DBLP:conf/eccv/ChenZPSA18} & ResNeSt-101 & 46.9& 66M & 1051G \\
DeepLab.v3+~\cite{DBLP:conf/eccv/ChenZPSA18} & ResNeSt-200 & 48.4 & 88M & 1381G  \\
SETR~\cite{zheng2021rethinking} & T-Large & 50.3 & 308M & -  \\\hline
UperNet~\cite{DBLP:conf/eccv/XiaoLZJS18} & ResNet-101 & 44.9 & 86M & 1029G \\
UperNet & DeiT-S & 44.0 & 52M & 1099G \\
UperNet & Swin-T & 46.1 & 60M & 945G  \\
UperNet & Swin-T$^*$ & 44.5 & 60M & 945G  \\
\hline
UperNet & LG-T & 45.3 & 64M & 957G \\
\hline
\end{tabular}
\end{center}
    \caption{Results on the ADE20K val set. $*$ indicates the results from the public codes of Swin-Transformer~\cite{DBLP:journals/corr/abs-2103-14030}. Our models are implemented based on this code base.}
    \label{tab:seg}
    \normalsize
\end{table*}

\subsection{Semantic Segmentation}

\textbf{Experimental Settings}. We conduct experiments of semantic segmentation on the ADE20K dataset~\cite{DBLP:journals/ijcv/ZhouZPXFBT19}, which includes 150 classes, and 20K, 2K images for training, and validation, respectively. We follow most of the settings as~\cite{DBLP:journals/corr/abs-2103-14030}. We resize and crop images to $512\times512$ resolutions for training, and resize images to $2048\times512$ for evaluation. Our model is pretrained on ImageNet-1K. AdamW with $\beta_1=0.9$ and $\beta_2=0.999$ is used as optimizer, and weight decay is set to 0.01. Our model is trained for 160K iterations. The initial learning rate is $6\times10^{-5}$ using a poly scheduler for learning rate decay. Batch size is set to 2 on each GPU. It takes near 1 day to train UperNet with LG-T, on a cluster with 8 V100 GPUs. We use the same training augmentation methods as~\cite{DBLP:journals/corr/abs-2103-14030}, and no test-time augmentation is employed.

\textbf{Results Analysis}. We apply our model to semantic segmentation by combining LG-T with UperNet~\cite{DBLP:conf/eccv/XiaoLZJS18}. As shown in Table~\ref{tab:seg}, our model achieves improvements compared with ResNet-101 ($\uparrow$0.4\%), DeiT-S ($\uparrow$1.3\%) and Swin-T$^*$ ($\uparrow$0.8\%), with a similar number of parameters and computational complexity. Although our model achieves inferior performance compared with state-of-the-art segmentation models, \emph{i.e.}, DeepLab V3 + ResNeSt-200 and SETR, these models require a larger model size and more computation. From this perspective, our model achieves a good trade-off between performance and efficiency.

\section{Conclusion}
In this paper, we propose to incorporate local and global attention into a Vision Transformer. By building a multi-path structure in the hierarchical Vision Transformer framework, our model takes advantage of both local the feature learning mechanisms in CNNs and global feature learning mechanisms in Transformers. We conduct thorough studies on two computer vision tasks, and the results demonstrate that our framework yields improved performance with limited sacrifice in model parameters and computational overhead. As a multi-path framework, one of the limitations of our model is that the inference speed is lower than its single-path Transformer counterparts. We will try to improve the efficiency in our future work.

\bibliographystyle{unsrt}
\bibliography{egbib}



\end{document}